\pdfoutput=1

\documentclass[11pt]{article}

\usepackage{acl}

\usepackage{times}
\usepackage{graphicx}
\usepackage{latexsym}
\usepackage{authblk}
\usepackage{soul}
\usepackage{hyperref}
\usepackage[utf8]{inputenc}
\usepackage{caption}
\usepackage{amsmath}
\usepackage{amsthm}
\usepackage{booktabs}
\usepackage{enumitem}
\usepackage{multirow}
\usepackage{algpseudocode}
\usepackage{algorithmicx}
\usepackage{algorithm}
\usepackage{IEEEtrantools}
\usepackage{amssymb, amscd, amsfonts}
\usepackage{subfigure}
\usepackage{arydshln}
\usepackage{makecell}
\usepackage{siunitx}
\usepackage{xspace,mfirstuc,tabulary}
\usepackage{microtype}
\usepackage{bm}
\usepackage{multicol}
\usepackage{xcolor}
\usepackage[switch]{lineno}

\newcommand{\unire}{\textsc{UniRE}\xspace}
\newcommand{\mymodel}{\textsc{HioRE}\xspace}

\definecolor{red}{RGB}{217, 38, 38}
\definecolor{green}{RGB}{0, 153, 0}

\makeatletter
\def\adl@drawiv#1#2#3{%
        \hskip.5\tabcolsep
        \xleaders#3{#2.5\@tempdimb #1{1}#2.5\@tempdimb}%
                #2\z@ plus1fil minus1fil\relax
        \hskip.5\tabcolsep}
\newcommand{\cdashlinelr}[1]{%
  \noalign{\vskip\aboverulesep
           \global\let\@dashdrawstore\adl@draw
           \global\let\adl@draw\adl@drawiv}
  \cdashline{#1}
  \noalign{\global\let\adl@draw\@dashdrawstore
           \vskip\belowrulesep}}
\makeatother

\newcommand\blfootnote[1]{%
  \begingroup
  \renewcommand\thefootnote{}\footnote{#1}%
  \addtocounter{footnote}{-1}%
  \endgroup
}

\usepackage[T1]{fontenc}

\usepackage{microtype}

\title{\mymodel: Leveraging High-order Interactions for Unified Entity Relation Extraction}

\author[1]{\bf Yijun Wang}
\author[4]{\bf Changzhi Sun$^\dagger$}
\author[3]{\bf Yuanbin Wu}
\author[5]{\bf Lei Li}
\author[2]{\bf Junchi Yan}
\author[4]{\bf Hao Zhou}
\affil[1]{WeChat Technical Architecture Department, Tencent Inc.}
\affil[2]{Department of Computer Science and Engineering, Shanghai Jiao Tong University}
\affil[3]{School of Computer Science and Technology, East China Normal University}
\affil[4]{ByteDance AI Lab} 
\affil[5]{University of California Santa Barbara}
\affil[  ]{\tt \{yijunwang.cs,czsun.cs\}@gmail.com}

\begin{document}

\maketitle

\begin{abstract}
    Entity relation extraction consists of two sub-tasks: entity recognition and relation extraction.
    Existing methods either tackle these two tasks separately or unify them with word-by-word interactions.
    In this paper, we propose \mymodel, a new method for unified entity relation extraction.
    The key insight is to leverage the \textit{high-order interactions}, i.e., the complex association among word pairs, which contains richer information than the first-order word-by-word interactions.
    For this purpose, we first devise a W-shape DNN (WNet) to capture coarse-level high-order connections.
    Then, we build a heuristic high-order graph and further calibrate the representations with a graph neural network (GNN).
    Experiments on three benchmarks (ACE04, ACE05, SciERC) show that \mymodel achieves the state-of-the-art performance on relation extraction 
    and an improvement of $1.1\sim1.8$ F1 points over the prior best unified model.
\end{abstract}

\blfootnote{$^\dagger$Corresponding Author.}

\section{Introduction}
\label{sec:intro}
\begin{figure}[tb!]
    \begin{center}
        \includegraphics[height=1.8in]{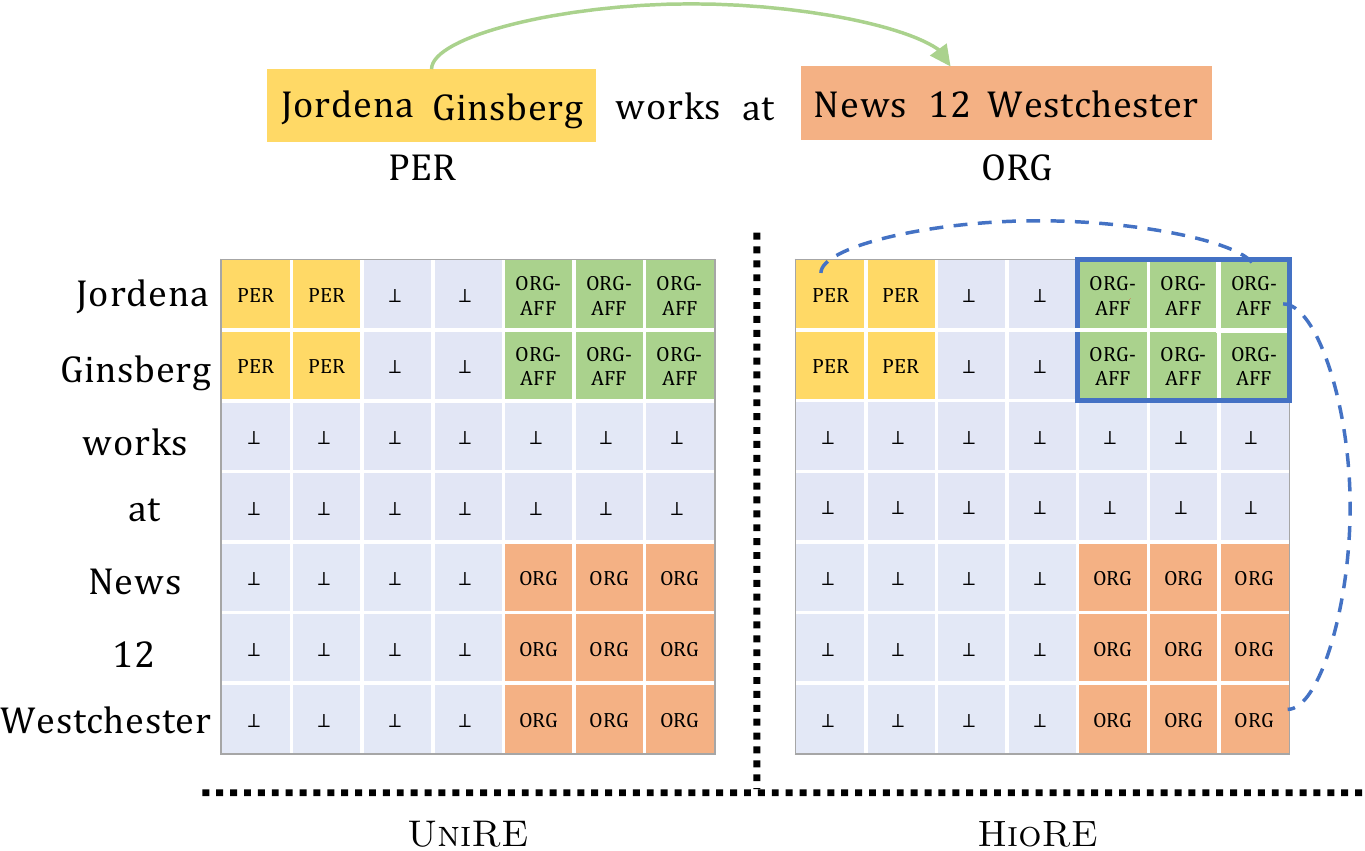}
    \end{center}
    \vspace{-10pt}
    \caption{Example of high-order features for unified entity relation extraction. Each cell in the table corresponds to a word pair in the sentence. Entities are squares on the diagonal, and relations are rectangles off the diagonal. Unlike \unire, which only considers first-order word-by-word interactions and fills each cell independently, \mymodel leverages high-order interactions between word pairs to improve unified entity relation extraction, e.g., the pairs (Jordena, Jordena), (Westchester, Westchester) provide a rich context for tagging the pair (Jordena, Westchester). In the table, the cells belong to the same blue rectangle and the cells connected by the dashed lines provide much complementary information for each other.
    }
    \label{fig:task}
    \vspace{-10pt}
\end{figure}

Automatically extracting entities and relations from the free text is a fundamental task of NLP.
It aims to identify typed spans (\emph{entities}) and assign a semantic relation for each entity pair (\emph{relations}).
As shown in Fig.~\ref{fig:task},
a person (\texttt{PER}) entity ``Jordena Ginsberg'' and an organization (\texttt{ORG}) entity ``News 12 Westchester'' have an affiliation (\texttt{ORG-AFF}) relation.\footnote{In Fig.~\ref{fig:task}, a special symbol $\bot$ indicates that there is no semantic relation.}

Most mainstream approaches fall into one of two classes: \textit{pipeline} or \textit{joint}. 
Pipeline approaches use two independent models to predict entities and relations, respectively, while joint approaches build connections between the two sub-models by parameters sharing~\cite{miwa-bansal-2016-end,katiyar-cardie-2017-going,sanh2019hierarchical,luan2019general,wang-lu-2020-two} or joint decoding~\cite{yang-cardie-2013-joint,li-ji-2014-incremental,katiyar2016investigating,zhang2017end,sun-etal-2018-extracting,sun-etal-2019-joint,lin2020joint}.
Due to adopting two separate label spaces for entity and relation, most joint methods can only learn shallow interactions between the two sub-models, leaving the deep interactions unexplored.
To address this issue, a recent joint paradigm \unire~\cite{wang2021unire} proposes a unified label space and tackle the two sub-tasks (i.e., entity recognition and relation extraction)  by a single model, achieving the state-of-the-art performance with faster speed.

The key insight of \unire is leveraging word-by-word interactions to construct a 2D table, where each cell corresponds to a word pair in the sentence.
In this way, the two sub-tasks are unified into one table-filling problem.
However, considering that it only encodes the first-order interactions between two words with the biaffine attention mechanism~\cite{DozatM17}, we wonder whether there exist high-order interactions that may further advance unified entity relation extraction.
Revisiting the 2D table, we find that the connections among multiple cells provide more complementary information than an isolated cell.
For example in Fig.~\ref{fig:task}, neighboring cells (marked by the blue rectangle) potentially share the same label and can provide references for each other.
Besides, cells across different regions (marked by dashed lines) imply some specific constraints on multi-entity, multi-relation, and entity-relation, e.g., the \texttt{ORG-AFF} relation usually indicates the existence of \texttt{PER} and \texttt{ORG} entity.
Here the connections among multiple word pairs (cells) are built on top of the first-order word-by-word interactions and hence are regarded as high-order interactions.
To this end, we attempt to leverage the high-order interactions for unified entity relation extraction.

In this paper, we follow \unire to formulate entity relation extraction as a table filling problem and propose \mymodel, which incorporates high-order interactions in a coarse-to-fine manner.
We first devise a W-shape DNN (WNet), which aggregates the first-order (word-by-word) representations around local neighbors to model the coarse-level high-order interactions among word pairs.
Note that we also utilize the attention information from pre-trained language models in the WNet, which provide additional guidance concerning the word-by-word similarity.
After that, we define a heuristic high-order graph upon the table to learn fine-grained high-order information.
Specifically, we assign a node for each cell and design two edge-linking strategies: the static graph and the dynamic graph.
The static graph is defined according to hand-crafted principles, which build connections among some specific cells, i.e., the diagonal cells and cells in the same row (column).
The dynamic graph is learned with a binary classification task, which further prunes some redundant edges.
Based on the heuristic graph, we adopt a graph neural network (GNN) to calibrate the representation of each node and bring global constraints into the final decision.

We summarize our contributions as follows:
\begin{itemize}[leftmargin=*, itemindent=0.6pc]
    \item We propose \mymodel for unified entity relation extraction, which leverages high-order interactions among word pairs.
    We first develop a WNet to learn coarse-level information. Then we employ a GNN on top of the heuristic high-order graph to calibrate the final representations.
    \item Experiments on three benchmarks (ACE04, ACE05, and SciERC) demonstrate the effectiveness of the high-order information and show that \mymodel achieves state-of-the-art relation performance on the three benchmarks.
\end{itemize}

\section{Background}
\label{sec:background}
Given an input sentence $s = \{ w_i \}_{i=1} ^ n$ of length $n$ ($w_i$ is a word),
it aims to extract an entity set $\mathcal{E}$ and a relation set $\mathcal{R}$.
An entity $e \in \mathcal{E}$ is a text span $\{ w_i\}_{i \in e}$ \footnote{$i \in e$ denotes $\mathtt{start}(e) \leq i \leq \mathtt{end}(e)$, where $\mathtt{start}(e)$ is the start word offset of $e$ in the sentence $s$ and $\mathtt{end}(e)$ is similar.} with an pre-defined entity type $y_e \in \mathcal{Y}_e$, e.g., person (\texttt{PER}), organization (\texttt{ORG}).
A relation $r \in \mathcal{R}$ is a triplet $(e_1, e_2, y_r)$,
where $e_1, e_2$ are two entities and $y_r \in \mathcal{Y}_r$ is a pre-defined relation type
describing the semantic relation among two entities, e.g., organization affiliation relation (\texttt{ORG-AFF}).
$\mathcal{Y}_e, \mathcal{Y}_r$ denote the set of possible entity types and relation types, respectively.

\unire formulates the entity relation extraction as a table filling task.
For the input sentence $s$, it maintains a 2D table $\mathcal{T}^{n \times n}$, where each cell is assigned a label $y_{i, j} \in \mathcal{Y}$ ($ \mathcal{Y} = \mathcal{Y}_e \cup \mathcal{Y}_r \cup \{\bot\} $, $\bot$ denotes no relation).
For each entity $e$, the corresponding cells $y_{i, j} (i \in e, j \in e)$ are filled with $y_e$.\footnote{We do not consider nested entities in this paper.}
For each relation $(e_1, e_2, y_r)$, the corresponding cells $y_{i, j}(i \in e_1, j \in e_2)$ are filled with $y_r$.
Lastly, the remaining cells are filled with $\bot$.

Next, we will briefly describe overall \unire architecture in four parts:
the sentence encoder, the table encoder, table filling, and table decoding.
Due to space limitations, please refer to \cite{wang2021unire} for more details.

\paragraph{Sentence Encoder}
For the input sentence $s$,
\unire adopts a pre-trained language model (PLM) like BERT \cite{devlin-etal-2019-bert} as the sentence encoder to obtain contextual representations.
The output is calculated via
\begin{IEEEeqnarray*}{c}
    \{\bm{h}_1, \dots, \bm{h}_{n}\} = \mathtt{PLM}(\{\bm{x}_1, \dots, \bm{x}_{n}\}),
\end{IEEEeqnarray*}
where $\bm{x}_i$ sums corresponding word, position, and segmentation embeddings of word $w_i$.
To better model the directional information of word-by-word interactions, a head multi-layer perceptron (MLP) and a tail MLP are applied on each $\bm{h}_i$:
\begin{IEEEeqnarray*}{c}
    \bm{h}_i^\mathrm{head} = \mathtt{MLP}_\mathrm{head}(\bm{h}_i), ~ ~
    \bm{h}_i^\mathrm{tail} = \mathtt{MLP}_\mathrm{tail}(\bm{h}_i).
\end{IEEEeqnarray*}

\paragraph{Table Encoder}
\label{bg:table}
\unire adopts the deep biaffine attention mechanism~\cite{DozatM17} to obtain the scoring vector $\bm{g}_{i, j} \in \mathbb{R}^{|\mathcal{Y}|}$ of each cell $(i, j)$ in the table $\mathcal{T}$:
\begin{IEEEeqnarray*}{c}
    \bm{g}_{i, j} = \mathtt{Biaff}(\bm{h}_i^\mathrm{head}, \bm{h}_j^\mathrm{tail}).
\end{IEEEeqnarray*}

\paragraph{Table Filling}
With the scoring vector $\bm{g}_{i, j}$, 
\unire yields a categorical probability distribution over the unified label space $\mathcal{Y}$ by the softmax function.
Given the gold label $y_{i, j}^*$, the objective function is to minimize
\begin{IEEEeqnarray*}{c}
    \mathcal{L}_\mathrm{entry} = - \frac{1}{n^2} \sum_{i, j} \log p(\bm{y}_{i, j} = y_{i, j}^* | s), \\
        p(\bm{y}_{i, j}|s) = \mathtt{Softmax}(\bm{g}_{i, j}).
\end{IEEEeqnarray*}
In the training phase, it imposes two structural constraints (symmetry objective $\mathcal{L}_\mathrm{sym}$ and implication objective $\mathcal{L}_\mathrm{imp}$) and jointly optimized them with $\mathcal{L}_\mathrm{entry}$.
To some extent, these two constraints can be viewed as artificial high-order features. 
In this work, we further investigate the high-order information in both coarse and fine level.

\paragraph{Table Decoding}
In the testing phase, \unire proposes an approximate decoding algorithm to compute the final extraction results.
The main idea is to firstly decode spans, then decode the entity label of each span, and lastly decode the relation label of each entity pair.

\section{\mymodel}
\label{sec:approach}
\begin{figure*}[t]
    \begin{center}
        \includegraphics[width=6in]{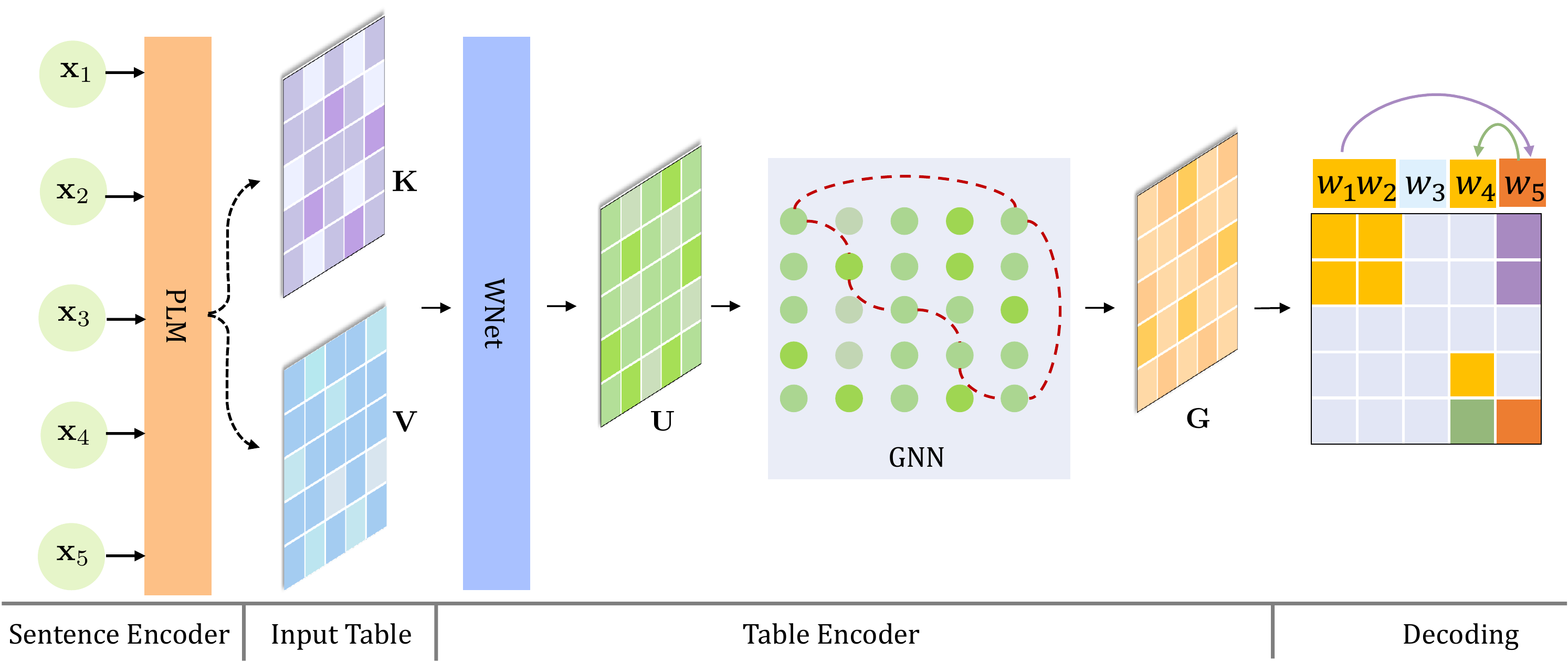}
    \end{center}
    \vspace{-10pt}
    \caption{Overview of \mymodel. 
    $\bm{K}$ denotes the similarity matrix between any two words (i.e., attention matrix from PLM). 
    $\bm{V}$ denotes the vanilla table representation of word pairs.
    $\bm{U}$ is a representation that incorporates the interactions among word pairs in coarse-level.
    $\bm{G}$ is the calibrated representation through GNN, which leverages the fine-grained high-order interactions.
    }
    \label{fig:model}
    \vspace{-10pt}
\end{figure*}

Following the unified paradigm \unire~\cite{wang2021unire}, we propose \mymodel and formulate the entity relation extraction as a table filling problem.
Unlike \unire that considers only the first-order word-by-word interactions, \mymodel further leverages the high-order information among word-pair in a coarse-to-fine manner.
The schematic description of \mymodel is shown in Fig.~\ref{fig:model}.
With the vanilla table representations of word pairs, we first devise a W-shape convolutional network to aggregate the coarse-level high-order information among local neighbors (Section~\ref{sec:local-info}).
After that, we build a heuristic high-order graph and use a GNN to propagate and calibrate the fine-grained information (Section~\ref{sec:graph-info}).
Next, we will explain the proposed model in detail.
\subsection{Coarse-level High-order Information}
\label{sec:local-info}
Unlike \unire that processes each cell (representing a word pair) independently when computing the table representation, we desire to aggregate the neighboring information within the table.
For this purpose, we treat the 2D table as an image and apply convolution to capture the neighboring information.
For the network design, we find that the U-shaped convolutional network~ \cite{RFB15a} combines the encoder-decoder architecture with skip connections, which facilitates the extraction of both structural and semantic information.
Inspired by the classical U-shape architecture, we devise a W-shape (i.e., double U-shape) convolutional network (WNet) to capture the coarse-level high-order information. 
Next, we will detail the construction of the image-like input and the WNet.



\paragraph{Input Table Construction}
On top of the contextual representation of each word in the sentence $s$,
 we construct two types of input tables to encode word-to-word information.
\begin{itemize}[leftmargin=*, itemindent=0.6pc]
    \item For each cell $(i, j)$ of the table $\mathcal{T}$, 
we construct a vanilla cell representation $\bm{v}_{i, j}$ based on the word representations $\bm{h}_i^\mathrm{head}$ and $\bm{h}_j^\mathrm{tail}$ as
\begin{IEEEeqnarray*}{c}
    \bm{V}_{i, j} = [\bm{h}_i^\mathrm{head}; \bm{h}_j^\mathrm{tail}; \bm{h}_i^\mathrm{head} - \bm{h}_j^\mathrm{tail}; \bm{h}_i^\mathrm{head} \odot \bm{h}_j^\mathrm{tail}; \bm{c}_{|i-j|}],
\end{IEEEeqnarray*}
where $\bm{c}_{|i-j|} $ is the distance embedding,
$\odot$ denotes element-wise multiplication 
and $[\cdot;\cdot]$ is the concatenation operation.
We denote the input matrix as $\mathbf{V} \in \mathbb{R}^{n \times n \times v}$ and $\bm{V}_{i, j} \in \mathbb{R}^v$ ($v = 750$).

\item 
Besides, pre-trained language models (PLM) provide not only the contextual representations for each token but also the attention scores that reflect the word-to-word similarity. To take advantage of the attention information, we extract the attention matrix from PLM as the table input $\bm{K}^{n \times n \times k}$,
which is collected from all heads of each layer (e.g., $k = 12 \times 12$ for the BERT base model) in the pretrained model.
\footnote{We take the first sub-word for each token to collect attention scores. Since the number of attention matrix channels is too large for ALBERT-xxlarge model (k = 768), we omit the attention matrix input in ALBERT-xxlarge-based model for efficiency.}
\end{itemize}

\begin{figure}[t]
    \begin{center}
        \includegraphics[width=2.6in]{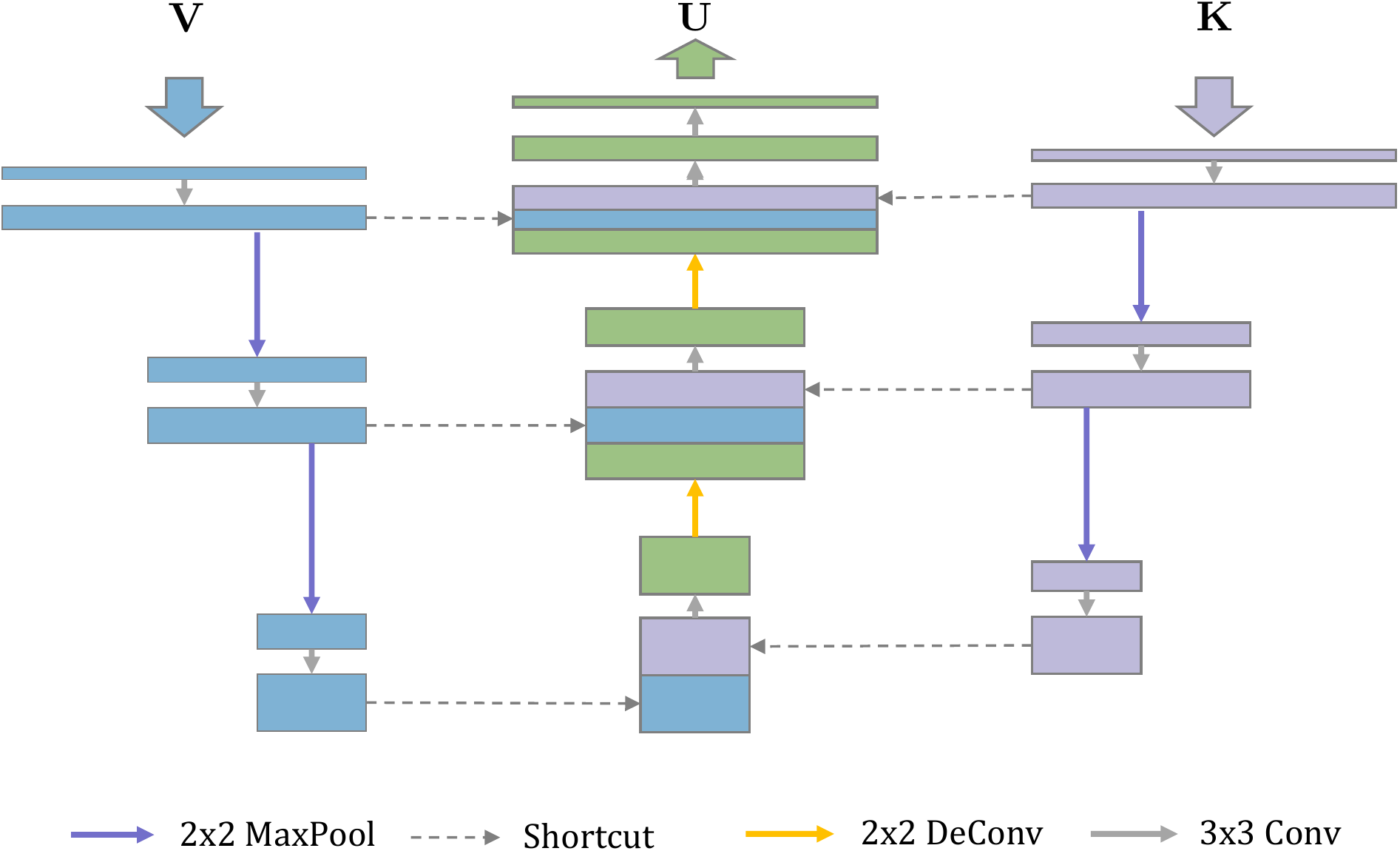}
    \end{center}
    \vspace{-5pt}
    \caption{Overview of WNet. It consists of two encoders (left and right) and one decoder (middle) with shortcut connections. The two encoders extract multi-scale representations for the two input tables ($\bm{V}$ and $\bm{K}$) , then the decoder output the table $\bm{U}$ that incorporates multi-scale high-order information.}
    \label{fig:unet}
    \vspace{-10pt}
\end{figure}
\paragraph{WNet}
Given the word-by-word representation $\bm{V}$ and the attention signal $\bm{K}$, we carefully devise a W-shape DNN, namely WNet, to aggregate the high-order interactions among the neighboring word pairs.
WNet is a variant of UNet that follows an encoder-decoder schema, with multiple shortcuts connecting the encoded features to the decoding process.
Unlike the conventional UNet that consists of one encoder and one decoder, WNet contains two encoders to process V and K separately but share one decoder.
As shown in Fig.~\ref{fig:unet}, the two types of information are compacted by the corresponding encoder and are lately integrated during the decoding process.
In this way, the output $\bm{U}\in \mathbb{R}^{n \times n \times u}$ summarizes both the neighboring high-order interactions and word-by-word correlations.
\begin{IEEEeqnarray*}{c}
    \bm{U} = \mathtt{WNet}(\bm{V}, \bm{K}).
\end{IEEEeqnarray*}
\subsection{Fine-grained High-order Information}
\label{sec:graph-info}
After the coarse-level representation $\bm{U}$ aggregates the high-order interactions among neighboring cells, we desire to further incorporate the fine-grained information of entity-to-entity, relation-to-relation, and entity-to-relation.
For this purpose, we build a heuristic graph $\mathcal{G}$ upon the 2D table, with each cell $(i,j)$ corresponding to a graph node and $\bm{U}_{i,j}$ as the initial node embedding. 
Thus, the total number of nodes in the graph $\mathcal{G}$ is $n^2$.
Next, we establish the cell-to-cell connections as the edges between two nodes.
The edges in the graph $\mathcal{G}$ are linked by two strategies, resulting the \textit{static graph} and the \textit{dynamic graph}.

\paragraph{Static Graph}
The static graph is defined by the following two hand-crafted principles:
\begin{itemize}[leftmargin=*, itemindent=0.6pc]
    \item We connect any two cells on the diagonal, namely, $(i, i)$ and $(j, j)$ where $i \neq j$.
    This type of edge may reveal the interactions between entities and advance entity prediction with more other entities' features.
    \item For each off-diagonal cell $(i, j)$,  we connect it to its two corresponding diagonal cells $(i, i)$ and $(j, j)$.
    This type of edge may reveal the interactions between entities and relations. For example, the edges between the pair (Jordena, Westchester) and (Jordena, Jordena), (Westchester, Westchester) in Fig.~\ref{fig:task}, which may help (Jordena, Westchester) learn the information from its argument entities, and vice versa.
\end{itemize}
Besides, the two principles can also build connections between relations by multi-hopping, providing other relations' feature for relation prediction.

This strategy uses the diagonal cells as pivots and outputs a sparse graph.
Nevertheless, the static graph possibly contain some redundant edges:
\begin{itemize}[leftmargin=*, itemindent=0.6pc]
    \item If the label of a diagonal cell $(i, i)$ is $\bot$, the edges connecting $(i, i)$ to other diagonal cells $(j, j)$ are redundant edges.
    \item If the label of an off-diagonal cell $(i, j)$ is $\bot$, the edges connecting $(i, j)$ to $(i, i)$ and $(i, j)$ to $(j, j)$ are redundant edges.
\end{itemize}
To prune these redundant edges, we further investigate the dynamic graph.

\paragraph{Dynamic Graph}
We introduce a binary classification task to predict whether the label of each cell is $\bot$.
Given the initial embedding $\bm{U}_{i,j}$ for the cell $(i,j)$,
we adopt a classifier to compute the probability of the binary label $\bm{b}_{i, j}$.
\begin{IEEEeqnarray*}{c}
    p(\bm{b}_{i,j} | s) = \mathtt{Softmax}(\mathbf{W}_\mathrm{bin} \bm{U}_{i,j}),
\end{IEEEeqnarray*}
where $\mathbf{W}_\mathrm{bin}$ is a learnable parameter.
The training objective is to minimize
\begin{IEEEeqnarray*}{c}
    \mathcal{L}_\mathrm{bin}=-\frac{1}{n^2}\sum_{i,j} \log p(\bm{b}_{i,j} = b_{i, j}^*|s),
\end{IEEEeqnarray*}
where the gold binary label $b_{i,j}^*$ are transformed from the table annotations $y_{i,j}^*$.
\footnote{$b_{i,j}^* = 1$ if $y_{i,j}^* \neq \bot$, and $b_{i,j}^* = 0$ if $y_{i,j}^* = \bot$.}
According to the predicted label, we build the edges of the dynamic graph by the following rules:
\begin{itemize}[leftmargin=*, itemindent=0.6pc]
    \item If $\hat{b}_{i,j} = 1$ where $i \neq j$, we connect $(i, j)$ to $(i, i)$ and $(i, j)$ to $(j, j)$;
    \item If $\hat{b}_{i,i} = \hat{b}_{j,j} = 1$ where $i \neq j$, we connect $(i, i)$ to $(j, j)$. 
\end{itemize}

In practice, we adopt either the static or the dynamic strategy to build a heuristic high-order graph $\mathcal{G}$, with the node embedding initialized by $\bm{U}$.
After that, a multi-layer Graph Neural Network (GNN) is used to obtain the final representation $\bm{G}$:
\begin{IEEEeqnarray*}{c}
\bm{G} = \mathtt{GNN}({\bm{U}},\mathcal{G}) \in \mathbb{R}^{n \times n \times d}.
\end{IEEEeqnarray*}


\subsection{Training and Decoding}
\label{sec:train-infer}
To predict each label of the table $\mathcal{T}$,
the cell representation $\bm{G}_{i, j} $ is fed into the softmax function, yielding a categorical probability distribution over the unified label space $\mathcal{Y}$ as
\begin{IEEEeqnarray*}{c}
    p(\bm{y}_{i, j} | s) = \mathtt{Softmax}(
    \mathbf{W}_y \bm{G}_{i,j}),
\end{IEEEeqnarray*}
where $\mathbf{W}_y$ is a learnable parameter.
Given the input sentence $s$ and the gold label of each cell $y_{i,j}^*$,
the training objective is to minimize
\begin{IEEEeqnarray*}{rrl}
    \mathcal{L}_\mathrm{entry} &=&- \frac{1}{n^2}\sum_{i, j}{\log p(\bm{y}_{i, j} = y_{i,j}^*|s)}.
\end{IEEEeqnarray*}
With the static graph, we only optimize $\mathcal{L}_\mathrm{entry}$.
And with the dynamic graph, we jointly optimize $\mathcal{L}_\mathrm{entry}$ and binary classification objective $\mathcal{L}_\mathrm{bin}$ as $\mathcal{L}_\mathrm{entry} + \mathcal{L}_\mathrm{bin}$.

In the decoding phase, we use the same algorithm as \unire to extract all entities and relations based on table predictions.


\section{Experiments}
\label{sec:exp}
\paragraph{Datasets}

We evaluate our model on three common used datasets,
i.e., ACE04 \citep{doddington-etal-2004-automatic},
ACE05 \citep{walker2006ace},
and SciERC \citep{luan-etal-2018-multi}.
We use the same data splits and pre-processing as \citep{li-ji-2014-incremental, miwa-bansal-2016-end, wang2021unire}.

\paragraph{Evaluation}
Following prior works, we report \textbf{F1} score with \textbf{Micro-averaging} strategy for performance comparison.
Specifically, i) we consider the entity is correctly labeled if its type and boundaries match those of a gold entity, and ii) the relation is correctly labeled when its type and two-argument entities (including the entity type and boundaries) both match those of a gold relation, i.e., the \textbf{Strict Evaluation} criterion.

\paragraph{Experimental Settings}
We set the hidden size of head/tail MLP as $d=150$ and the activation function as GELU.
Following \cite{wang2021unire}, the batch size is 32, the learning rate is 5e-5 with weight decay 1e-5, and the optimizer is AdamW with $\beta_1 = 0.9$ and $\beta_2 = 0.9$.
We train the model with a maximum of 300 epochs and employ an early stop strategy. 

We tune all hyper-parameters on the development set of ACE05, then directly use the same hyper-parameters on ACE04 and SciERC.
For each experiment, we run our model 3 runs and report the averaged scores.
We pick the best model based on the averaged F1 score of the entity and relation on the development set.
All experiments are conducted on an NVIDIA Tesla V100 GPU (32G).

\subsection{Performance Comparison}
We compare the performance of our model with previous methods in Table~\ref{tab:main-results}.
In general, our model achieves the best relation performance on three benchmarks with different pre-trained models. 
Specifically, with the base encoder (BERT$_{\texttt{BASE}}$ and SciBERT), \mymodel increases the relation performance by +1.7 (ACE04), +1.0 (ACE05), +1.4 (SciERC) absolute F1 points over the previous best results. 

Note that though \mymodel is slightly inferior to the previous best model in the entity detection, it significantly advances both entity and relation performances upon \unire.
Specifically, \mymodel improves the relation performance by $1.1\sim1.8$ F1 points on three datasets and entity performance by 0.5 and 0.4 points on ACE04 and ACE05, respectively.
These results demonstrate that \mymodel effectively mines the high-order information among word pairs.

\begin{table}[tb!]
    \centering
    \footnotesize
    \resizebox{0.48\textwidth}{!}{
    \begin{tabular}{cllcc}
        \toprule
        \textbf{Dataset} & \textbf{Encoder} & \textbf{Model}  & \textbf{Entity F1} & \textbf{Relation F1}              \\
        \midrule
        \multirow{10}{*}{\textbf{ACE04}}   &  \multirow{2}{*}{LSTM} & M\&W(\citeyear{miwa-bansal-2016-end})  & 81.8  & 48.4  \\
        ~ & ~ & k\&C(\citeyear{katiyar-cardie-2017-going})  & 79.6  & 45.7  \\
        \cmidrule(lr){2-5}
        ~ & BERT$_{\texttt{LARGE}}$ & Li(\citeyear{li-etal-2019-entity})  & 83.6  & 49.4  \\
        \cmidrule(lr){2-5}
        ~ & \multirow{3}{*}{BERT$_{\texttt{BASE}}$} & Z\&C(\citeyear{zhong2021frustratingly})\textsuperscript{$\diamond$}              & \textbf{89.2}  & 60.1          \\
        ~ & ~ & \unire(\citeyear{wang2021unire})\textsuperscript{$\diamond$}           & 87.7         & 60.0          \\
        \cmidrule(lr){3-5}
        ~ & ~ & \mymodel\textsuperscript{$\diamond$}  & 87.2(-0.5) &  \textbf{61.8}\textcolor{green}{(+1.8)}  \\
        \cmidrule(lr){2-5}
        ~ & \multirow{4}{*}{ALBERT$_{\texttt{XXLARGE}}$}   & W\&L(\citeyear{wang-lu-2020-two})  & 88.6   & 59.6          \\
        ~ & ~ & Z\&C(\citeyear{zhong2021frustratingly})\textsuperscript{$\diamond$} & \textbf{90.3} & 62.2          \\
        ~ & ~ & \unire(\citeyear{wang2021unire})\textsuperscript{$\diamond$}      & 89.5 & 63.0 \\
        \cmidrule(lr){3-5}
        ~ & ~ & \mymodel\textsuperscript{$\diamond$} & 90.0\textcolor{green}{(+0.5)}  & \textbf{64.2}\textcolor{green}{(+1.2)} \\
        \midrule
        \multirow{12}{*}{\textbf{ACE05}}  &  \multirow{3}{*}{LSTM} & M\&W(\citeyear{miwa-bansal-2016-end})  & 83.4  & 55.6  \\
        ~ & ~ & k\&C(\citeyear{katiyar-cardie-2017-going})  & 82.6  & 53.6  \\
        ~ & ~ & Sun(\citeyear{sun-etal-2019-joint})  & 84.2  & 59.1  \\
        \cmidrule(lr){2-5}
        ~ & BERT$_{\texttt{LARGE}}$ & Li(\citeyear{li-etal-2019-entity})  & 84.8  & 60.2  \\
        \cmidrule(lr){2-5}
        ~ & \multirow{4}{*}{BERT$_{\texttt{BASE}}$}  & Wang(\citeyear{wang-etal-2020-pre}) & 87.2  & 63.2          \\
        ~ & ~ & Z\&C(\citeyear{zhong2021frustratingly})\textsuperscript{$\diamond$}           & \textbf{90.1}     & 64.8          \\
        ~ & ~ & \unire(\citeyear{wang2021unire})\textsuperscript{$\diamond$}                          & 88.8          & 64.3          \\
        \cmidrule(lr){3-5}
        ~ & ~ & \mymodel\textsuperscript{$\diamond$} & 89.6\textcolor{green}{(+0.8)} & \textbf{65.8}\textcolor{green}{(+1.5)}          \\
        \cmidrule(lr){2-5}
        ~ & \multirow{4}{*}{ALBERT$_{\texttt{XXLARGE}}$} & W\&L(\citeyear{wang-lu-2020-two})  & 89.5  & 64.3          \\
        ~ & ~ & Z\&C(\citeyear{zhong2021frustratingly})\textsuperscript{$\diamond$} & \textbf{90.9} & 67.0 \\
        ~ & ~ & \unire(\citeyear{wang2021unire})\textsuperscript{$\diamond$}  & 90.2      & 66.0          \\
        \cmidrule(lr){3-5}
        ~ & ~ & \mymodel\textsuperscript{$\diamond$}    & 90.6(+0.4)    & \textbf{67.1}\textcolor{green}{(+1.1)}    \\
        \midrule
        \multirow{4}{*}{\textbf{SciERC}} & \multirow{4}{*}{SciBERT} & Wang(\citeyear{wang-etal-2020-pre}) & 68.0  & 34.6          \\
        ~ & ~ & Z\&C(\citeyear{zhong2021frustratingly})\textsuperscript{$\diamond$} & \textbf{68.9} & 36.8                        \\
        ~ & ~ & \unire(\citeyear{wang2021unire})\textsuperscript{$\diamond$}  & 68.4 & 36.9 \\
        \cmidrule(lr){3-5}
        ~ & ~ & \mymodel\textsuperscript{$\diamond$}& 68.2(-0.2) & \textbf{38.3}\textcolor{green}{(+1.4)} \\
        \bottomrule
    \end{tabular}}
    \vspace{-5pt}
    \caption{Overall results on ACE04, ACE05, and SciERC. Numbers in the brackets are the gaps to the \textsc{UniRE}, which is the baseline model. $\diamond$ means that the model leverages cross-sentence context information. In green are the gaps of at least \textcolor{green}{+0.5} points.} 
    \label{tab:main-results}
    \vspace{-10pt}
\end{table}

\subsection{Ablation Study}
\label{sec:ablation}
\begin{table}[t]
    \centering
    \footnotesize
    \resizebox{0.48\textwidth}{!}{
        \begin{tabular}{lcccc}
            \toprule
            \multirow{2}{*}{\textbf{Settings}} & \multicolumn{2}{c}{\textbf{ACE05}} & \multicolumn{2}{c}{\textbf{SciERC}}                                 \\
            ~   & Ent F1            & Rel F1        & Ent F1           & Rel F1          \\
            \midrule
            Default     & \textbf{89.6}         & \textbf{65.8}                       & \textbf{68.2} & \textbf{38.3}          \\
            \midrule
            w/o $\bm{K}$             & 89.3                    & 65.2                                & 67.5          & 38.1 \\
            w/o WNet                  & 88.3                               & 63.3                                & 66.5          & 33.1          \\
            w/o GNN         & 89.0                               & 64.5                                & 67.0          & 37.4          \\
            \midrule
            dynamic graph &  89.2                               & 64.8                                & 66.9          & 37.5          \\
            biaffine attention & 88.8 & 65.0 & 67.2 & 36.4  \\
            constrained objectives & 89.4 & 65.5 & 67.6 & 35.9  \\
            \midrule
            single UNet                     & \textbf{89.6}                               & 65.6                       & 66.3 & 34.6         \\ 
            two separate UNets               & 88.8                              & 63.8                       & 67.9 & 37.6          \\
            \bottomrule
        \end{tabular}
    }
    \vspace{-5pt}
    \caption{Entity and relation performances under different settings on ACE05 and SciERC test sets. ``w/o'' means without. ``dynamic graph'' means that replacing static graph with dynamic graph. ``biaffine attention'' means that using biaffine attention to construct $\bm{V}$.  ``constrained objectives'' means that applying the symmetry and implication constrains introduced by \textsc{UniRE}. ``single UNet'' means that concatenating $\bm{V}$ and $\bm{K}$ then feed it into a single UNet. ``two separate UNets'' means that using two separate UNets for $\bm{V}$ and $\bm{K}$ inputs. 
    }
    \label{tab:ablation}
    \vspace{-10pt}
\end{table}
In this section, we evaluate the different components of \mymodel on ACE05 and SciERC.
From the results in Table~\ref{tab:ablation},
we have the following observations:
\begin{itemize}[leftmargin=*, itemindent=1pc]
    \item When the attention matrix input $\bm{K}$ is removed (line 2), both entity and relation performances decline on two datasets. It implies that the word-to-word similarity captured by the attention mechanism from the pre-trained language models helps extract entities and relations. 
    \item When we remove the WNet (line 3), the performance drops sharply. Specifically, entity F1 scores decrease by 1.3 points and 1.7 points, and relation F1 scores decrease by 2.5 points and 5.2 points. It shows that the WNet plays an essential role in \mymodel, which effectively utilizes coarse-level high-order information to provide powerful representations for the subsequent GNN encoder.
    \item When the GNN is removed (line 4), the performance also degenerates significantly (1.3 points and 0.9 points for relation scores), showing that the fine-grained interactions captured by the GNN further enhance the table representation. 
    \item Comparing with the ``Default'' setting, the ``dynamic graph'' achieves inferior performance (line 5), indicating that the dynamic graph is likely too sparse to capture sufficient interactions. And it requires more computation cost than the static graph, thus we use the static graph in the final model.
    \item We also try some tricks of \unire. For example,  we use the biaffine attention mechanism to construct the input table $\bm{V}$ (line 6) and apply the additional symmetry and implication constraints in the training phase (line 7). However, both tricks do not further advance the performance, even hazarding the relation performance on SciERC, which verifies that \mymodel has captured more effective high-order interactions.
    \item For the architecture of WNet, , we try some alternatives, i.e., a single UNet (line 8) and two separate UNets (line 9).
    When replacing the WNet with the UNet and concatenating $\bm{V}$ and $\bm{K}$ as one input, the performance on ACE05 is comparable but that on SciERC drops sharply.
    Besides, adopting two separate UNets for two input tables (i.e., $\bm{V}$ and $\bm{K}$) is also inferior to WNet.
    These results show that the WNet is more effective and stable on both datasets. 
\end{itemize}

\subsection{Impact of Different Settings for GNN}

\begin{figure}[t]
    \begin{center}
        \includegraphics[width=2.8in]{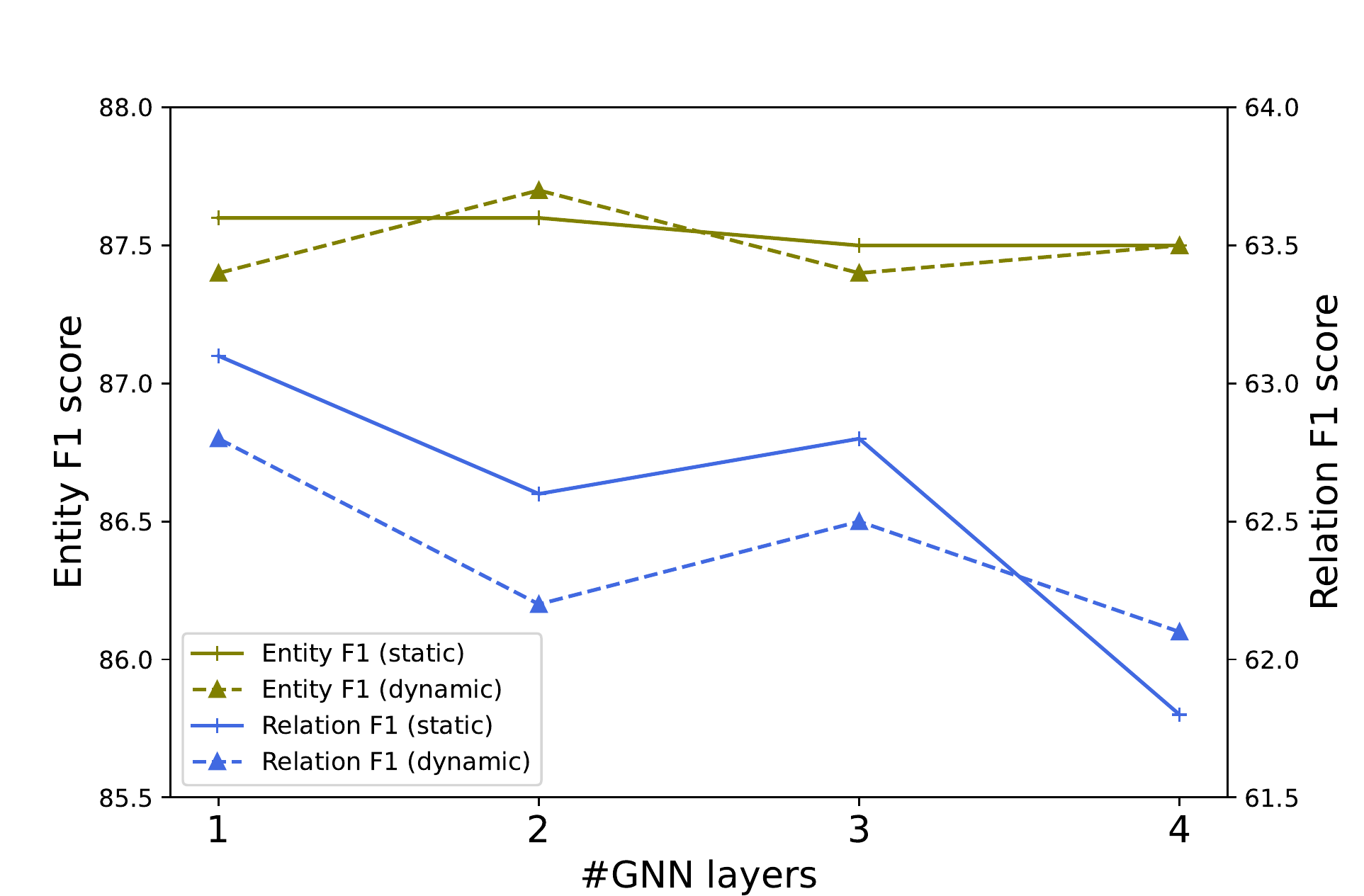}
    \end{center}
    \vspace{-10pt}
    \caption{Entity and relation performances with different numbers of GNN layers and different graph schemas (static and dynamic) on the ACE05 dev set. }
    \label{fig:gcn}
    \vspace{-10pt}
\end{figure}

We evaluate our model under different settings of GNN, i.e., different numbers of GNN layers and different graph schemas, on the development set of ACE05 and SciERC.
As shown in Fig.~\ref{fig:gcn}, when increasing the number of GNN layers from 1 to 4, the relation performance gradually decreases while the entity performance almost keeps still.
The result shows that the simplest 1-layer GNN is more effective, so we adopt 1-layer GNN in the final \mymodel design.
Next, the entity performance is still stable when changing the graph schema from the static graph to the dynamic graph. 
But the relation performance of the dynamic graph is slightly inferior to that of the static graph (-0.3 points).
We think that the dynamic graph introduces some errors in the binary relation classification procedure, which subsequently hazards the relation performance.



\subsection{Inference Speed}
\begin{table}[t]
    \centering
    \footnotesize
    \resizebox{0.48\textwidth}{!}{
        \begin{tabular}{lcccccc}
            \toprule
            \multirow{2}{*}{\textbf{Model}}                  & \multirow{2}{*}{\textbf{Parameters}}  & \multicolumn{2}{c}{\textbf{ACE05}} & \multicolumn{2}{c}{\textbf{SciERC}}                                  \\
            ~                                                                                 & ~                           & \makecell[c]{Rel                                                                                          \\(F1)}                                                         & \makecell[c]{Speed\\(sent/s)} &  \makecell[c]{Rel\\(F1)} & \makecell[c]{Speed\\(sent/s)} \\
            \midrule
            Z\&C(\citeyear{zhong2021frustratingly}) & 219M & 64.6 & 13.1 & 36.7 & 20.5 \\
            \unire(\citeyear{wang2021unire}) & 110M    & 64.3  & \textbf{200.5} & 36.9 & \textbf{205.4}  \\
            \textsc{UniRE+}(\citeyear{wang2021unire}) & 114M  & 63.9 & 190.1 & 36.1 & 198.5 \\
            \midrule
            \mymodel  & 134M    & \textbf{65.8}   & 136.4  & \textbf{38.3} & 133.5          \\
            dynamic graph & 134M &  64.8    & 49.5  & 37.5 & 36.5          \\
            \bottomrule
        \end{tabular}
    }
    \vspace{-5pt}
    \caption{Comparison of accuracy and efficiency on ACE05 and SciERC test sets with the \textbf{same} setting. \textsc{UniRE+} is a variant of \unire with larger hidden size of the biaffine model. Both \mymodel and ``dynamic graph'' adopt 1-layer GNN.}
    \label{tab:speed}
    \vspace{-10pt}
\end{table}
In this section, we evaluate the inference speed of our model on ACE05 and SciERC (Table~\ref{tab:speed}).
We use the same pre-trained language model and batch size as \cite{wang2021unire} and obtain the results of other models under the same machine configuration.
Compared with \unire, the inference speed of our model drops by 30\% due to the more complex network architecture.
But \mymodel achieves significantly superior relation performance (+1.5 and +1.4 absolute F1 score on ACE05 and SciERC), and we consider it an acceptable trade-off between performance and speed.
Besides, it is worth noting that the speed of \mymodel is still quite competitive when compared with prior work \cite{zhong2021frustratingly}.
Note that when adopting the dynamic graph in our model, the inference speed drops dramatically, i.e., a quarter of that of \textsc{UniRE} on ACE05, and one-eighth of that of \textsc{UniRE} on SciERC.
Therefore, we adopt the static graph in the final model, which strikes a good balance between accuracy and efficiency.

To study the impact of the number of model parameters, we tried to increase the number of UniRE's parameters by increasing the hidden size of the biaffine model (from 150 to 512), achieving \textsc{UniRE+} (Table~\ref{tab:speed}).
But the performance of UniRE+ did not improve further or even decreased on ACE05 and SciERC. 
Thus, we attribute the performance improvement of \mymodel mainly to the novel network architecture.
Moreover, compared with prior work \cite{zhong2021frustratingly}, \mymodel is still quite compact.




\subsection{Error Analysis}
\label{sec:error-analysis}
\begin{table*}[t]
    \centering
    \footnotesize
    \resizebox{0.95\textwidth}{!}{
        \begin{tabular}{lcl}
        \toprule
        \textbf{Type}  & \textbf{Model} & \textbf{Sentence} \\ 
        \midrule
        \multirow{3}{*}{IE}  & Gold & After a $\text{[police]}_{\texttt{ORG-AFF}}^{\texttt{ORG}}$  $\text{[officer]}_{\texttt{ORG-AFF}}^{\texttt{PER}}$ arrived , the second $\text{[bag]}_{}^{\texttt{WEA}}$ exploded , seriously injuring $\text{[him]}_{}^{\texttt{PER}}$ .
        \\ 
        \cmidrule(lr){2-3}
        ~ & \textsc{UniRE}   & After a $\text{[police]}_{\texttt{ORG-AFF}}^{\texttt{ORG}}$  $\text{[officer]}_{\texttt{ORG-AFF}}^{\texttt{PER}}$ arrived , the second \textcolor{red}{bag} exploded , seriously injuring $\text{[him]}_{}^{\texttt{PER}}$ .            
        \\ 
        \cmidrule(lr){2-3} 
        ~ & \mymodel & After a $\text{[police]}_{\texttt{ORG-AFF}}^{\texttt{ORG}}$  $\text{[officer]}_{\texttt{ORG-AFF}}^{\texttt{PER}}$ arrived , the second $\text{[bag]}_{}^{\texttt{WEA}}$ exploded , seriously injuring $\text{[him]}_{}^{\texttt{PER}}$ .                          
        \\ 
        \midrule
        \multirow{3}{*}{MR} & Gold & It had reached a deal with the $\text{[British]}_{\texttt{GEN-AFF}}^{\texttt{GPE}}$ $\text{[arm]}_{\texttt{GEN-AFF|PART-WHOLE}}^{\texttt{ORG}}$ of French distributors $\text{[Pathe]}_{\texttt{PART-WHOLE}}^{\texttt{ORG}}$ .
        \\ 
        \cmidrule(lr){2-3}
        ~ & \textsc{UniRE}  & It had reached a deal with the $\text{[British]}_{\texttt{GEN-AFF}}^{\texttt{GPE}}$ $\text{[arm]}_{\texttt{GEN-AFF|PART-WHOLE}}^{\texttt{ORG}}$ of French $\text{[\textcolor{red}{distributors}]}_{\texttt{PART-WHOLE}}^{\texttt{ORG}}$ \textcolor{red}{Pathe}.
        \\ 
        \cmidrule(lr){2-3}
        ~ & \mymodel & It had reached a deal with the $\text{[British]}_{\texttt{GEN-AFF}}^{\texttt{GPE}}$ $\text{[arm]}_{\texttt{GEN-AFF|PART-WHOLE}}^{\texttt{ORG}}$ of French distributors $\text{[Pathe]}_{\texttt{PART-WHOLE}}^{\texttt{ORG}}$ .
        \\ 
        \midrule
        \multirow{3}{*}{LDR} & Gold & $\text{[Protesters]}_{\texttt{PHYS-1|PHYS-2}}^{\texttt{PER}}$ also gathered in their thousands in $\text{[Halifax]}_{\texttt{PHYS-1}}^{\texttt{GPE}}$ , Edmonton and $\text{[Vancouver]}_{\texttt{PHYS-2}}^{\texttt{GPE}}$ .
        \\
        \cmidrule(lr){2-3}   
        ~ & \textsc{UniRE} & $\text{[\textcolor{red}{Protesters}]}_{\texttt{PHYS-1}}^{\texttt{PER}}$ also gathered in their thousands in $\text{[Halifax]}_{\texttt{PHYS-1}}^{\texttt{GPE}}$ , Edmonton and $\text{[\textcolor{red}{Vancouver}]}_{}^{\texttt{GPE}}$ .
        \\ 
        \cmidrule(lr){2-3}                                         
        ~ & \mymodel    & $\text{[Protesters]}_{\texttt{PHYS-1|PHYS-2}}^{\texttt{PER}}$ also gathered in their thousands in $\text{[Halifax]}_{\texttt{PHYS-1}}^{\texttt{GPE}}$ , Edmonton and $\text{[Vancouver]}_{\texttt{PHYS-2}}^{\texttt{GPE}}$ .
        \\ 
        \midrule
        \end{tabular}
        }
        \vspace{-5pt}
        \caption{Examples from the ACE05 test set annotated by \textsc{UniRE} and \mymodel for comparison. Words in red are \textcolor{red}{wrong} annotations.}
        \label{tab:err}
        \vspace{-20pt}
\end{table*}

\begin{figure}[t]
    \begin{center}
        \includegraphics[width=2.8in]{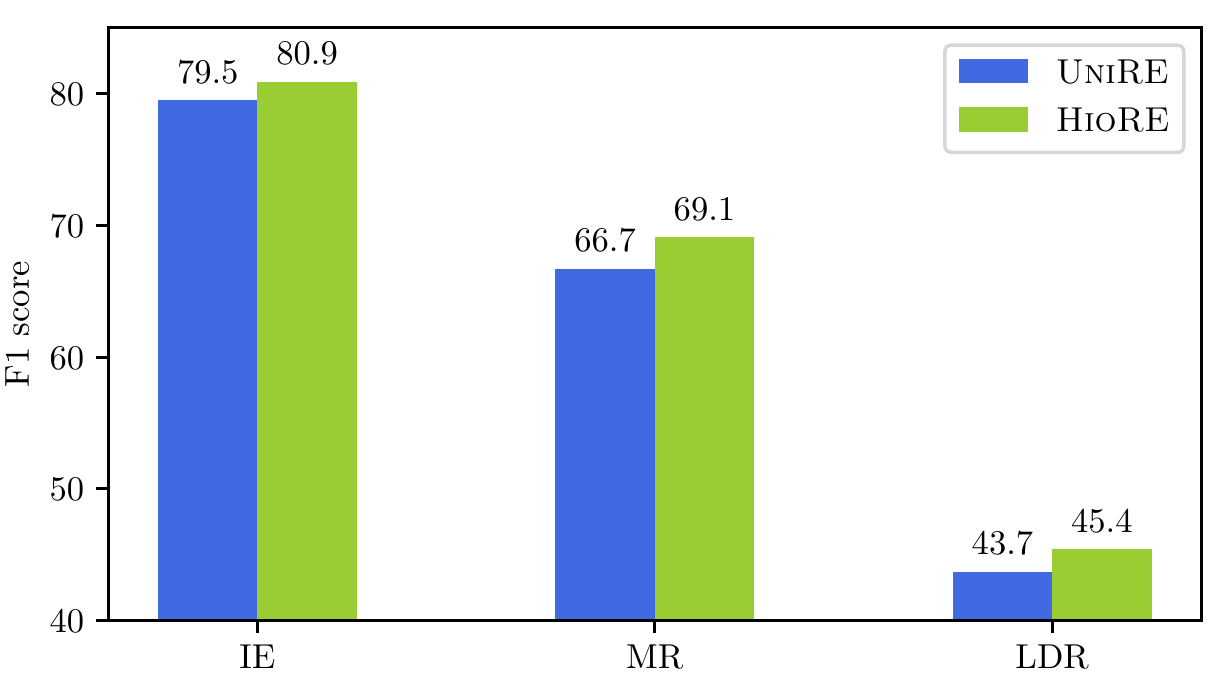}
    \end{center}
    \vspace{-10pt}
    \caption{Performance comparison in three hard situations on the ACE05 test set. 
    }
    \label{fig:error}
    \vspace{-15pt}
\end{figure}

In this section, we compare \mymodel with \textsc{UniRE} \cite{wang2021unire} in three hard situations (Fig.~\ref{fig:error}): 
i) \textit{isolated entity} (IE), i.e., an entity that does not participate in any relation; 
ii) \textit{multi-relation} (MR), that multiple relations exist in one sentence; 
and iii) \textit{long-distance relation} (LDR), that the distance between a relation's two argument entities exceeds 4.
Some specific examples of the typical error are presented in Table~\ref{tab:err}.
\begin{itemize}[leftmargin=*, itemindent=1pc]
    \item 
    In the ``IE'' case, \mymodel achieves better entity performance (+1.4 F1 score) than \textsc{UniRE}.
    An example is that \mymodel detects the isolated \texttt{WEA} entity ``bag'' while \textsc{UniRE} does not.
    It shows that with entity connection on the table graph, \mymodel can capture more inter-entity interactions and detect more semantic information.
    Though outperforming \textsc{UniRE}, performance in the ``IE'' situation is significantly lower than that on the complete test set, raising a crucial problem for further exploration.
    \item 
    For relation extraction, \mymodel significantly outperforms \textsc{UniRE} on both ``MR'' (+ 2.4 points) and ``LDR'' (+1.7 points), showing that \mymodel makes the best of the interactions between multiple relations and effectively builds long-distance dependency between words.
    As shown in Table~\ref{tab:err}, \mymodel successfully extracts two overlapped relations (line 6), while \textsc{UniRE} fails to detect the \texttt{PART-WHOLE} relation (line 5).
    In addition, \textsc{UniRE} misses a longer distance relation \texttt{PHYS} (line 8) while \mymodel correctly recognizes it.
    On the whole, both \mymodel and \textsc{UniRE} can handle the ``MR'' situation but are not satisfying in the ``LDR'' case.
    The performance of detecting long-distance relations is only two-thirds of that on the complete test set, indicating that the long-distance relation problem is another challenge that waits to be solved.
\end{itemize}

\section{Related Work}
\label{sec:related}
\paragraph{Unified Entity Relation Extraction} 
is first introduced by \cite{zheng-etal-2017-joint}, which formulates entity relation extraction as a sequence labeling problem. 
Then \cite{ijcai2018-620} convert this task into a graph generation problem and solve it with a transition-based parsing system.
Besides, \cite{zeng-etal-2018-extracting,zhang-etal-2020-minimize,Nayak_Ng_2020} adopt a Seq2Seq model to generate relation triplets.
\cite{wang-etal-2020-tplinker} propose a token pair linking problem to extract relation triplets, which is also based on table filling.
Unlike thse work that do not model entity types, both \unire~\cite{wang2021unire} and our model \mymodel jointly model entity types and relation types and extract entity and relation in a unified label space.
Comparing with \unire, our model further considers more high-order interactions among word pairs.

\vspace{-2pt}
\paragraph{UNet}
is first used to biomedical image segmentation and gradually becomes a popular backbone in the image segmentation.
Interestingly, recent work \cite{liu-etal-2020-incomplete,ijcai2021-551} try to introduce UNet into some NLP tasks, e.g., \cite{liu-etal-2020-incomplete} successfully tackle the incomplete utterance rewriting with the UNet.
Thus, we also borrow ideas from UNet to learn high-order interactions.

\vspace{-2pt}
\paragraph{Graph Neural Network} 
has been widely used in many NLP tasks,
e.g., semantic role labeling \cite{marcheggiani-titov:2017:EMNLP2017}, machine translation \cite{bastings-etal-2017-graph}, relation extraction \cite{zhang-etal-2018-graph,sun-etal-2019-joint}.
\cite{sun-etal-2019-joint} first introduce GNN into joint entity relation extraction, 
which regards each entity and relation as node and constructs a bipartite graph for joint type inference.


\section{Conclusion}
In this paper, we focus on leveraging high-order interactions to advance unified entity relation extraction.
We capture coarse-level high-order information with the WNet and adopt a GNN on the heuristic graph to further calibrate representations.
Experimental results show that our model \mymodel achieves state-of-the-art performance on three benchmarks.
The unified extraction for overlapped entities we leave for future work.

\bibliography{main}

\end{document}